\documentclass[journal]{IEEEtran}

\usepackage{authblk}
\usepackage[space]{cite}
\usepackage{filecontents}
\usepackage{graphicx}
\usepackage{caption}
\usepackage{hyperref}

\usepackage[T1]{fontenc}
\usepackage{etoolbox}

\makeatletter
\patchcmd{\maketitle}{\@fnsymbol}{\@alph}{}{}  
\makeatother

\begin{document}

\title{Neuroimaging Modality Fusion in Alzheimer's Classification Using Convolutional Neural Networks}

\author[1]{Arjun Punjabi}
\author[2,3]{Adam Martersteck}
\author[1]{Yanran Wang}
\author[2]{Todd B. Parrish}
\author[1]{Aggelos K. Katsaggelos}
\author[ ]{the Alzheimer's Disease Neuroimaging Initiative}

\affil[1]{Electrical Engineering and Computer Science, McCormick School of Engineering, Northwestern University}
\affil[2]{Department of Radiology, Feinberg School of Medicine, Northwestern University}
\affil[3]{Mesulam Cognitive Neurology and Alzheimer's Disease Center, Northwestern University}

\maketitle

\begin{abstract}
Automated methods for Alzheimer's disease (AD) classification have the potential for great clinical benefits and may provide insight for combating the disease. Machine learning, and more specifically deep neural networks, have been shown to have great efficacy in this domain. These algorithms often use neurological imaging data such as MRI and PET, but a comprehensive and balanced comparison of these modalities has not been performed. In order to accurately determine the relative strength of each imaging variant, this work performs a comparison study in the context of Alzheimer's dementia classification using the Alzheimer's Disease Neuroimaging Initiative (ADNI)\footnote{Data used in preparation of this article were obtained from the Alzheimer's Disease Neuroimaging Initiative (ADNI) database (adni.loni.usc.edu). As such, the investigators within the ADNI contributed to the design and implementation of ADNI and/or provided data but did not participate in analysis or writing of this report. A complete listing of ADNI investigators can be found at: \url{http://adni.loni.usc.edu/wp-content/uploads/how_to_apply/ADNI_Acknowledgement_List.pdf}} dataset. Furthermore, this work analyzes the benefits of using both modalities in a fusion setting and discusses how these data types may be leveraged in future AD studies using deep learning.
\end{abstract}


\section{Introduction}
\IEEEPARstart{A}{lzheimer's} disease (AD) is a neurodegenerative disorder characterized by cognitive decline and dementia. The number of individuals living with AD in the United States is expected to reach 10 million by the year 2025 \cite{hebert2004state}. As a result, automated methods for computer aided diagnosis could greatly improve the ability to screen at-risk individuals.

Such methods typically take as input patient data including demographics, medical history, genetic sequencing, and neurological images among others. The resulting output is health status indicated by a diagnosis label, which may also include a probabilistic uncertainty on the prediction. This particular investigation will focus on two different neuroimaging modalities: structural T1-weighted MRI and AV-45 amyloid PET. The primary goal of the investigation is to compare the efficacy of each of these modalities in isolation as well as when both are used as simultaneous input to a fusion system.

The algorithmic design of these methods can vary, but recent successes in machine learning have opened the floodgates for a plethora of deep neural networks trained for computer aided diagnosis. Given the visual nature of the input data, this work opted to apply a model well suited for computer vision tasks: the convolutional neural network (CNN). The following sections  will focus on related approaches to the AD classification problem, the methodology of both the network and data processing pipeline, and a discussion of the classification results.

\section{Related Work}
Computer aided diagnosis methods in this domain have spanned the gamut of algorithmic design. Earlier methods often applied linear classifiers like support vector machines (SVM) to hand-crafted biological features \cite{cuingnet2011automatic}. These features can be defined at the individual voxel level, as in the case for tissue probability maps, or at the regional level, including cortical thickness and hippocampal shape or volume. The 2011 comparison performed in \cite{cuingnet2011automatic} found that whole brain methods generally achieved higher classification accuracy than their region-based counterparts. Additionally, there was evidence to suggest that certain data pre-processing methods, namely the DARTEL registration package \cite{ashburner2007fast}, can substantially impact classification results. These two findings informed the decision to use whole brain volumes in this work and design a robust registration pipeline before the classification algorithm.

Similar linear classifier or SVM-based methods exist that align with these ideas. In \cite{kloppel2008automatic}, gray matter tissue maps were classified with an SVM. A more complex scheme exists in \cite{liu2016inherent}, where template selection was performed on gray matter density maps and these features were clustered in preparation for SVM classification. As previously discussed, regional features can also be used as input to an SVM, such as spherical harmonic coefficients calculated from the hippocampus \cite{gerardin2009multidimensional}. In \cite{sabuncu2015clinical}, the analysis is extended to other linear classifiers, primarily comparing the performance between SVMs and variations of random forest classifiers on a large conglomerate of Alzheimer's datasets. These models can also extend to multiple data modalities as in \cite{zu2016label}, where features from MRI and PET data were extracted and combined with a kernel-based approach. In \cite{zhu2014novel}, the procedure was modified with a custom loss function in order to perform both diagnosis classification and cognitive score regression simultaneously using a modified support vector-based model trained with MRI, PET, and cerebrospinal fluid (CSF) images.

Despite the initial popularity of SVMs and linear classifiers, there has been a transition in the last several years toward more non-linear approaches. Namely, the introduction of artificial neural networks has transformed the landscape of automated Alzheimer's dementia diagnosis. However, even these methods have varied in construction. The works in \cite{suk2013deep, suk2014hierarchical} used a deep Boltzmann machine (DBM) to extract features from MRI and PET data which are then classified using an SVM. Similarly, a DBM was also used in \cite{li2015robust} to extract features from MRI and PET, but additionally included CSF and cognitive test scores. The features are still classified with an SVM. A more standard fully-connected neural network was trained on MRI images in \cite{yang2016spatial}, but performance was improved by adding spatial neighborhood regularization similar to the receptive field of convolutional kernels.

This leads to the current preferred machine learning model, the CNN. These models are well suited to tasks with 2D or 3D data due to the shared filter weights within each convolutional layer. A CNN was proposed in \cite{sarraf2016classification} that takes fMRI slices as input to a modified LeNet-5 CNN architecture \cite{lecun1998gradient}. The DeepAD paper \cite{sarraf2016deepad} further developed this notion by utilizing the more complex GoogleNet CNN \cite{szegedy2015going}. In \cite{li2014deep}, MRI and PET data were used to train a multimodal CNN for classification, but it also allowed for missing modalities and modality completion. Some methods opted to use autoencoders \cite{hinton1994autoencoders} which can employ convolutional filters, but structurally differ from CNNs. While CNNs are trained to map input images to some given representation, autoencoders are trained to perform dimensionality reduction and reconstruct the input image. In this manner, the features learned in the middle layer of an autoencoder can be extracted and classified with either linear or non-linear methods. In \cite{liu2015multimodal}, features from MRI and PET images were extracted using a stacked autoencoder which were then classified with softmax regression. On the other hand, the work in \cite{gupta2013natural} used an autoencoder on 2D MRI slices to learn basis features that are then used as CNN filters. A similar procedure was performed in \cite{payan2015predicting} that compared the performance between both 2D and 3D systems. An autoencoder was used in \cite{hosseini2016alzheimer} on full 3D MRI images to pre-train the layers of a CNN model, and this was expanded in \cite{vu2017multimodal}  to include the PET modality. 

Fundamentally, while methods exist that take advantage of multiple data types and apply state-of-the-art neural network architectures, comparison studies between modalities and architectures have been haphazard in their use of datasets and lacking in explanations of model efficacy. In some instances, subsets of larger databases were used without explanations of why certain images were included or excluded. Additionally, pre-processing pipelines differ between these various studies. Both of these factors contribute to incongruous modality comparison results between papers. Furthermore, the biological explanations for such discrepancies are often lacking or non-existent. This work is novel in both of these respects. First, the pre-processing used in this work is clearly explained and the rationale for each step is  provided. Second, the modality comparison results are discussed within a biological context that more effectively describes the relative performance of each data type.

\section{Methodology}
As previously alluded to in the discussion of related work, pre-processing operations can have a major impact on final classification performance. As a result, a pipeline was developed to correct several of the biases inherent in the imaging data. While the components of the pipeline employ existing algorithms, the overall structure differs from previous work and allows for a more fair comparison between the MRI and PET modalities.

This section also discusses the neural network architecture. The design of the network is similar to the CNN-based approaches discussed previously. Again, because the primary goal of the investigation is a comparison of data modalities rather than network styles, the CNN was designed to be representative of comparable methods comprised of standard network layers.

\subsection{Pre-processing}
The pre-processing pipeline aimed to correct several biases that can exist in raw MRI and PET data. This also removes the additional burden of the network learning methods to correct or overlook these biases. Instead, the network has the isolated task of finding patterns between healthy and Alzheimer's patients. The vast majority of related work also employs similar pre-processing techniques in order to combat standard problems; namely, most methods perform some kind of MRI bias field correction, volumetric skull stripping, and affine registration. This approach is nascent in its registration scheme in order to prepare data for longitudinal studies in addition to traditional single time instance analyses. This manifests itself in two ways. First, our current investigation that treats each of these scanning instances as distinct samples in the dataset is less biased by differences in pre-processing for each modality. Second, when the scanning instances are viewed jointly as a single sample in the dataset for a longitudinal study, the images are normalized both within the subject and among all subjects in the set. Future longitudinal studies that take advantage of this processed data will be discussed at the end of the paper. The building blocks of the pipeline are as follows:

\begin{figure*}
\centering
\includegraphics[width=\linewidth]{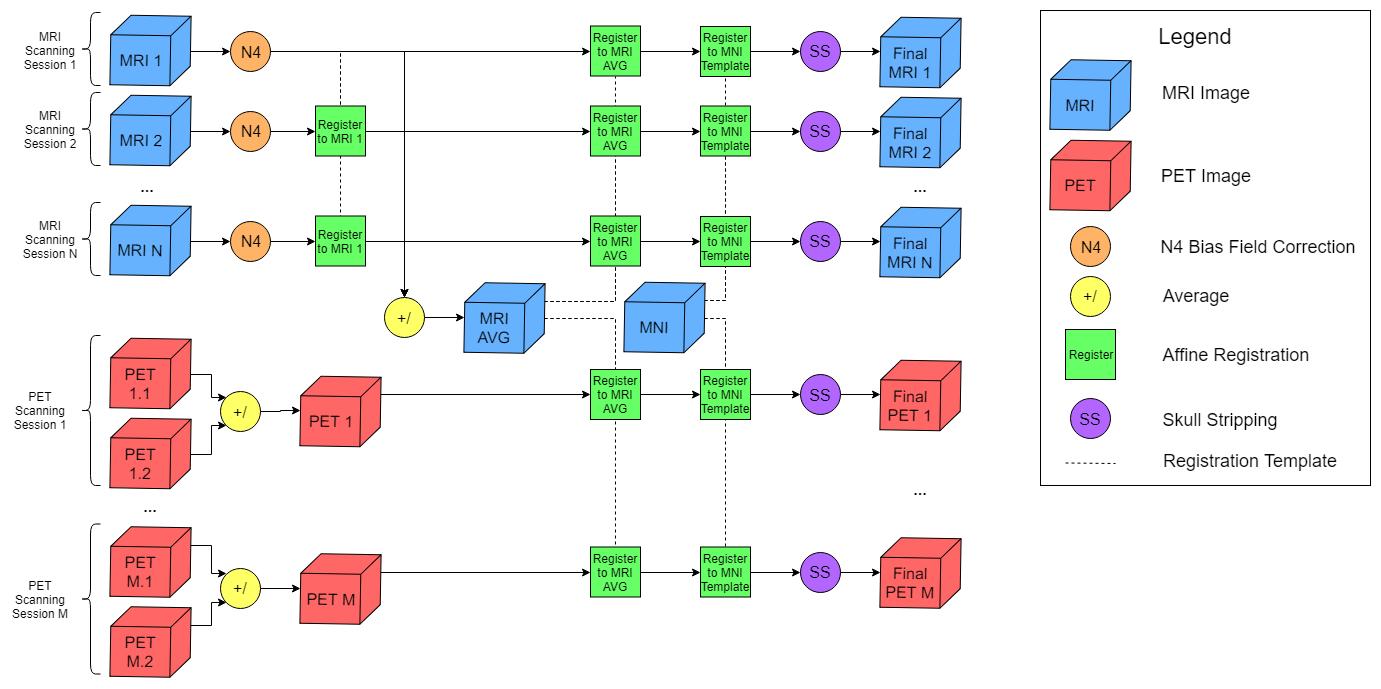}
\caption{Pre-processing pipeline for a single subject. A subject has N MRI scanning sessions and M PET scanning sessions; therefore, the pipeline yields N MRI images and M PET images. The pipeline is repeated for each subject in the dataset.}
\label{fig:pipeline}
\end{figure*}

\subsubsection{MRI Bias Field Correction}
MRI images can have a low frequency bias component as a result of transmit/receive inhomogeneities of the scanner \cite{mcveigh1986phase}. This spatial non-uniformity, while not always visually apparent, can cause problems for image processing pipelines. As a result, many MRI processing schemes begin by applying a bias field correction algorithm. Non-parametric non-uniform intensity normalization (N3) \cite{sled1998nonparametric} is a robust and well-established approach for removing this bias field. It optimizes for the slowly varying multiplicative field that, when removed, restores the high frequency components of the true signal. This work opted to employ a more recent update to this method known as N4 \cite{tustison2010n4itk}, which makes use of B-spline fitting for improved corrections. This step is performed on the raw MRI images and is unnecessary for the PET images.

\subsubsection{Affine Registration}
Both image modalities are registered using a linear affine transformation. Registration aims to remove any spatial discrepancies between individuals in the scanner, namely minor translations and rotations from a standard orientation. Typically, scans are registered to a brain atlas template, such as MNI152 \cite{fonov2011unbiased}. While this procedure is perfectly acceptable for traditional single time point analyses, this pipeline was designed to accommodate longitudinal studies as well. In such a setting, a patient in the dataset will have multiple scanning sessions at different times, but these images are aggregated and treated as a more complex representation of a single data point. As a result, it is beneficial to have congruence between the temporal scans in addition to registration to the standard template. Consequently, MRI and PET scans in the pipeline are registered first to an average template created from all MRI scans from a single patient, and then once more to the standard MNI152 space. The average template is created by registering all scans from one patient to a single scanning instance and then taking the mean of these images. Therefore, each subject will have unique average templates. Every MRI and PET scan is registered to the respective average template before the traditional registration with the MNI152 template. This ensures that all of the scans are registered both temporally within each patient's history and generally across the entire dataset. FSL FLIRT was used to perform the registrations \cite{jenkinson2002improved}.

\subsubsection{Skull Stripping}
Skull stripping is used to remove non-brain tissue voxels from the images. This is generally framed as a segmentation problem wherein clustering can be used to separate the voxels accordingly, as in FSL's brain extraction tool (BET) \cite{smith2002fast}. However, given that the scans were already registered to a standard space, skull stripping was a straightforward task. A brain mask in MNI152 space was used to zero out any non-brain voxels in both the MRI and PET images.

Figure \ref{fig:pipeline} shows the pipeline in its entirety. The process is performed for all MRI and PET images for a single patient in the dataset before proceeding to the next. N4 correction is applied to all of the MRI scans before any registration steps. All MRI scans are registered to the first scanning time point, and the resulting images are averaged to create the average template. The N4 corrected scans are registered to this space before being registered with the MNI152 template. The resulting images are then skull stripped using a binary mask.

Amyloid PET scans were collected over 20 minutes in dynamic list-mode 50 minutes post-injection of 370 MBq 18F-florbetapir. PET scans were attenuation corrected using a computed tomography scan. The first 10 minutes of PET acquisition was reconstruction into two 5 minute frames. Frames were motion corrected together and referenced (normalized) by the whole cerebellum. Each PET scan was registered to the individual's average T1 template with a 6 DOF registration and then the pre-computed 12 DOF registration from average T1 to MNI152 was concatenated and applied to the PET images to move them from native PET to MNI152 space. Finally, the PET images were skull stripped as above.

\begin{figure*}
\centering
\includegraphics[width=\linewidth]{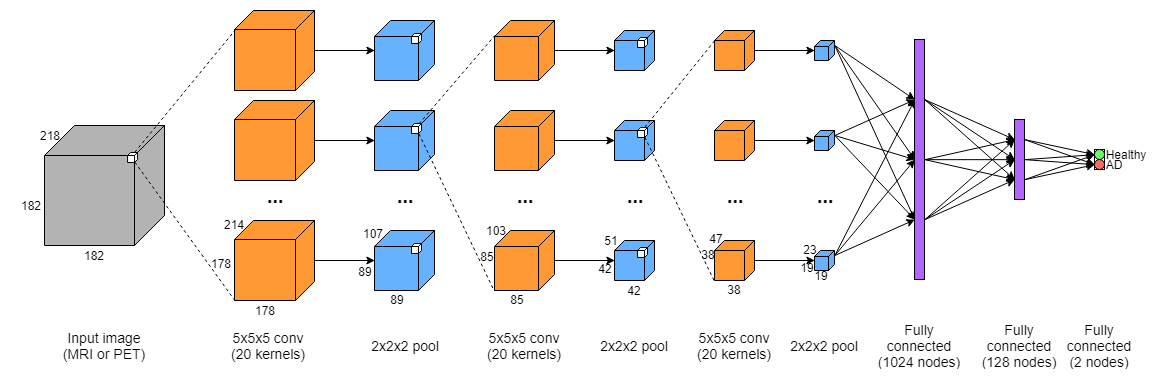}
\caption{Convolutional neural network for one modality. A single MRI or PET volume is taken as input, and the output is a binary diagnosis label of either "Healthy" or "AD".}
\label{fig:single_network}
\end{figure*}

\subsection{Network}
The CNN architecture is fairly traditional in its construction and is most similar to that in \cite{hosseini2016alzheimer}. The network takes as input a full 3D MRI or PET image and outputs a diagnosis label. While several processing layers exist in the network, there are only three different varieties: convolutional layers, max pooling layers, and fully connected layers. Convolutional layers constitute the backbone of the CNN. As the name suggests, 3D filters are convolved with the input to the layer. Each kernel is made of learned weights that are shared across the whole input image; yet, each processing layer can have multiple trainable kernels. This allows kernel specialization while still affording the ability to capture variations at each layer. Following convolutional layers, it is common to have max pooling layers. These layers downsample an input image by outputting the maximum response in a given region. For example, a max pooling layer with a kernel size of 2x2x2 will result in a output image that is half the input size in each dimension. Each voxel in the output will correspond to the maximum value of the input image in the associated 2x2x2 window. Fully connected layers are often placed at the end of a CNN. These layers take the region specific convolutional features learned earlier in the network and allow connections between every feature. The weights in these layers are also trainable; therefore, these layers aggregate the region features and learn global connections between them. As a result, the output of the final fully connected layer in the CNN is the final diagnosis label.

Figure \ref{fig:single_network} is a diagram of the final CNN architecture for a single modality. In this instance, the network accepts MRI or PET images of size 182x218x182 (due to the MNI template size), but in principle a CNN can accept an image of any size. The image is then processed by three pairs of alternating convolutional (20 kernels of size 5x5x5) and max pooling layers (kernel size 2x2x2). The convolutional layers use the ReLU \cite{hahnloser2000digital} activation function. Following these layers, the feature vector is flattened before being passed as input to a fully connected layer with 1024 nodes, a second fully connected layer of 128 nodes, and finally a fully connected layer with the number of diagnosis categories. In this case, there are 2 diagnosis categories corresponding to individuals with AD and healthy controls. The two fully connected layers also use the ReLU activation function, but the final classification is done with the softmax function.

\begin{figure*}
\centering
\includegraphics[width=\linewidth]{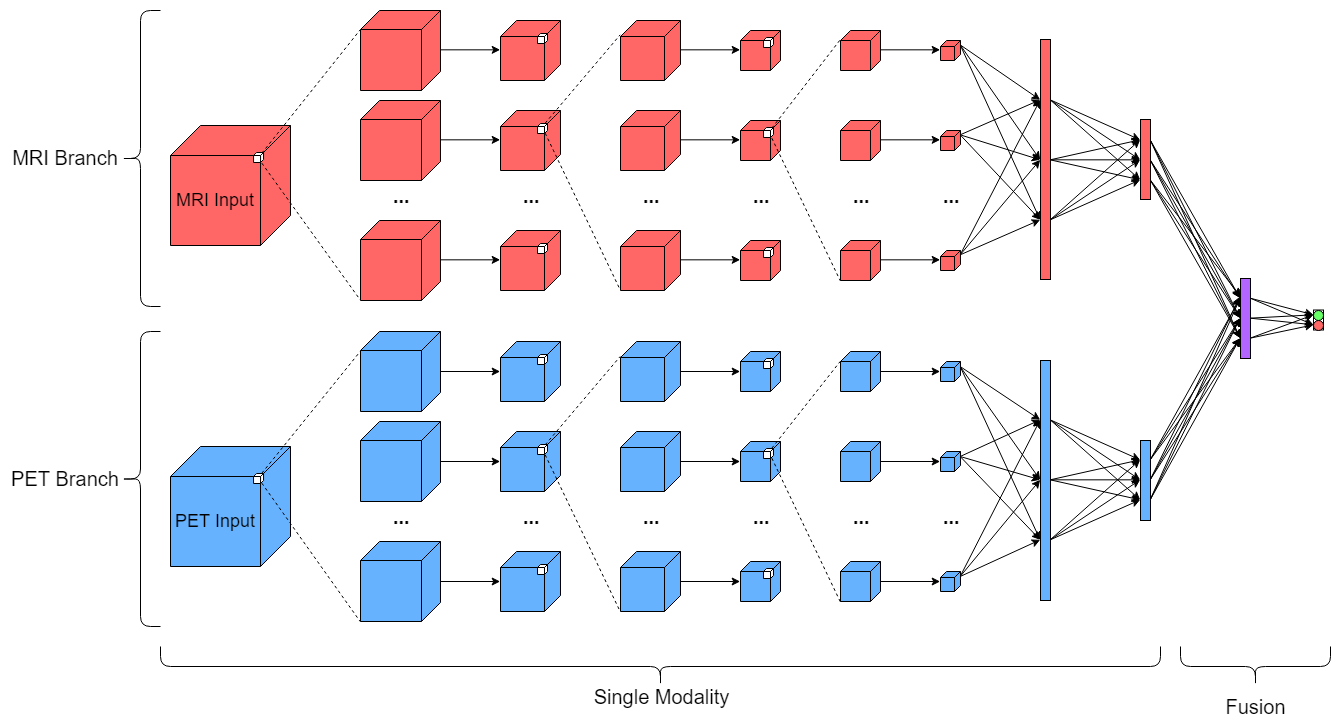}
\caption{Convolutional neural network for fusing MRI and PET modalities. An MRI and PET scan from a single patient is taken as input, and the output is again a binary diagnosis label.}
\label{fig:fusion_network}
\end{figure*}

Figure \ref{fig:fusion_network} shows the extension of the network for the fusion case. In this setting, the network takes both an MRI and PET image of size 182x218x182 as input into parallel branches. These branches are structured in the same manner as in the former case, but an additional fully connected layer of 128 nodes is added at the end in order to fuse the information from both modalities before the final classification is made. Additionally, the number of kernels in each convolutional layer was changed from 20 to 10 in order to keep the number of weights in the fusion network approximately the same as in the single modality network.

\section{Experimental Design}
Classification experiments were performed on the Alzheimer's Disease Neuroimaging Initiative (ADNI)\cite{jack2008alzheimer} database. The primary goal of ADNI has been to test whether serial MRI, PET, other biological markers, and clinical and neuropsychological assessment can be combined to measure the progression of mild cognitive impairment (MCI) and early Alzheimer's disease. The set has clinical data from hundreds of study participants including neuroimaging modalities, demographics, medical history, and genetic sequencing. This work analyzed T1-weighted MRI and amyloid PET images in addition to the diagnosis labels given to patients at each study visit. Neurological test scores were examined in order to validate these labels, but were not used during network training. Data was only used from participants who had at least one scanning session for both MRI and PET. Additionally, scanning sessions were not considered if neurological testing was not performed within 2 months of the scanning session. This was to ensure that the diagnosis label provided during the scanning sessions had clinical justification. As a result, a subset of 723 ADNI patients were used. As in \cite{sarraf2016deepad}, individual scanning sessions from the same patient were considered separately in this work. This resulted in 1299 MRI scans, each falling into either the healthy or AD category. Patients underwent less PET scanning, with a total of 585 scans. Classification experiments were initially performed using only one modality, either MRI or PET, and using the appropriate data subset. 

These sets were further split into training and testing components in order to ascertain the generalizability of the algorithm. When splitting the data into training and testing subsets, scanning sessions from a single patient were not used in both the testing and training subsets. In other words, all of a single patient's scans were used in one of the two subsets. This was done to ensure that the algorithm would not overfit to the patient's identity rather than learning the disease pattern. In some previous works, it is unclear whether this procedure was done. As a result, classification results in some previous work may have been inflated by models that overfit on certain individuals in the dataset.

Following this, fusion experiments were performed, where an MRI and PET scan from the same individual at a given time were used. Each scan was sent through parallel CNN branches. At the final fully connected layer of each branch, the features were merged into another fully connected layer that was used to produce the classification result. These experiments used the same number of data points as the PET experiments, albeit with each data point having an associated MRI and PET scan. Again, the testing and training subsets were made such that no patient's data was used in both subsets.

The neural network was constructed in Python using Keras \cite{chollet2015keras} as a front-end and Tensorflow \cite{abadi2016tensorflow} as the back-end deep learning framework. The optimization procedure used stochastic gradient descent with a learning rate of 0.0001 and a momentum of 0.9. Categorical cross-entropy was used to classify the results of the CNN into the diagnosis labels. Training was done on an Nvidia Titan Z GPU and took approximately 20 epochs to complete each experiment. Depending on the dataset size, epoch training times ranged between approximately 45 minutes and 1.5 hours.

\section{Results and Discussion}
Table \ref{table:1} details the results of the classification experiments. To reiterate, the structure of the MRI and PET networks are identical, as they both take in a single volume and have the same number of trainable weights. The fusion network takes in two volumes, one from each modality, into parallel branches that each have half the number of weights as a single MRI or PET network. Aside from a few extra weights at the end of the fusion network, the total number of weights in all three networks is roughly the same. Additionally, the fusion network used the same number of data points as the PET network, but each included two volumes instead of one. The MRI network was able to use more data points due to the larger number of MRI scanning sessions.

\begin{table}[h!]
\centering
\begin{tabular}{| c | c |}
\hline
Modality & Accuracy \\
\hline
MRI & 92\% \\
PET & 85\% \\
Fusion & 94\% \\
\hline
\end{tabular}
\caption{Classification Accuracies}
\label{table:1}
\end{table}

To begin, the MRI network is able to classify with 92\% accuracy. While this number is respectable, the performance could improve beyond 95\% by employing techniques such as those described in \cite{gupta2013natural, payan2015predicting, hosseini2016alzheimer}. However, the goal is again to compare the performance of the data modalities in the most balanced way possible. The inclusion of some of the more specific techniques in \cite{gupta2013natural, payan2015predicting, hosseini2016alzheimer}, such as pre-training the CNN filters with an autoencoder, does not enhance the modality comparison. Rather, the added complexity may obfuscate the findings if the pre-training effectiveness differed. As a result, the MRI performance of 92\% is used to compare to the PET and fusion performance.

In this vein, it can be seen that the PET network does not perform as well as the MRI network, as it only achieves 85\% accuracy. This result differs from \cite{vu2017multimodal}, which finds very small differences between MRI and PET versions of the same networks. This should be expected because the authors operate on FDG-PET, not the amyloid PET scans used in this work. Furthermore, the accuracy of the models in \cite{vu2017multimodal} is noticeably lower. Namely, the "Simple CNN" model, which is analogous to the models used in this work, performs at 80.62\% and 81.93\% for MRI and FDG-PET, respectively. Then, the authors' contribution of using a stacked autoencoder along with a CNN yielded accuracies of 85.24\% and 85.53\%, respectively. This again may point to the difficulty in comparing modality efficacies when the network structures become too complex. Fundamentally, this investigation more clearly illustrates the differences between the MRI and PET modalities and suggests that the MRI modality may be more informative to machine learning algorithms used for computer aided diagnosis.

In our experiments, the amyloid PET modality may have performed worse due to the temporal lag of amyloid accumulation and cognitive decline. In \cite{jack2013tracking}, the brain structure biomarkers (e.g., FDG-PET and structural MRI) much more closely follows the decline in cognition that a clinician uses to diagnose Alzheimer's dementia. Amyloid accumulation has been hypothesized to begin more than two decades before symptoms occur \cite{jack2013tracking}. In a longitudinal study of dominantly inherited Alzheimer's disease \cite{gordon2018spatial}, elevated amyloid PET signals were found 22 years before expected onset of symptoms.

Separate from the CNN pipeline, a standard method, previously described \cite{landau2012amyloid}, was used to calculate the total amyloid burden. Briefly, FreeSurfer \cite{fischl2012freesurfer,fischl2001automated} was used to parcellate the T1-weighted MRI scan taken closest to the amyloid PET visit. Whole cerebellar referenced cortical regions normalized by volume were used to calculate a single weighted standard uptake value ratio (SUVR). The previously defined cutoff of $\geq$  1.11 was used to define amyloid positivity \cite{landau2012amyloid}.

Out of the 11 amyloid PET scans that were incorrectly classified, 7 were controls and 4 were Alzheimer's dementia cases. All 7 control cases had elevated amyloid SUVR $\geq$ 1.11 (average SUVR 1.42 $\pm$ 0.12). Two Alzheimer's dementia cases were amyloid positive (i.e., true misclassification) and two Alzheimer's cases were amyloid negative (average SUVR 0.95 $\pm$ 0.03) and therefore are unlikely to have underlying Alzheimer's disease neuropathology. If the 7  elevated amyloid controls and 2 amyloid negative AD cases are removed, then the effective PET classification accuracy rises to 97\%. 

The newly proposed NIA-AA research criteria for Alzheimer's disease \cite{jack2018nia} points out that amnestic dementia diagnoses are not sensitive or specific for AD neuropathologic change. From 10 to 30\% of individuals classified as AD dementia do not display AD neuropathology at autopsy \cite{nelson2011alzheimer} and 30 to 40\% of individuals classified as unimpaired healthy have AD neuropathologic change at post-mortem examination \cite{bennett2006neuropathology, price1991distribution}. The proposed CNN here is capturing this mismatch between biomarker and diagnosis. The CNN labels healthy individuals with high amyloid PET as AD and those with Alzheimer's dementia and low amyloid PET as non-AD. Thus, while the phenomenon negatively impacts performance in this context, amyloid PET scans may be adept in a longitudinal study because elevated amyloid precedes symptom onset.

The final noteworthy result of the investigation is that the fusion network outperformed both the individual MRI and PET networks. Additionally, the fusion network outperformed the MRI network despite the fact that less data points were used. Having more PET scans available in the fusion case may further improve the accuracy. While the fusion performance is consistent with the results of \cite{vu2017multimodal, suk2014hierarchical}, this investigation more clearly demonstrates that even the less informative modality of PET can provide complementary information to improve classification accuracy. It is therefore most similar to the results of \cite{li2014deep}, but that work structures the task more as a modality completion problem rather than the more traditional classification task. Additionally, the work in \cite{li2014deep} used patches rather than total volumes. The full volume results obtained in the traditional CNN context are more illustrative of the phenomenon than the results of \cite{li2014deep}.

\section{Conclusion and Future Work}

This work compared the effectiveness of the MRI and amyloid PET modalities in the context of computer aided diagnosis using deep neural networks. Specifically, two identically structured CNNs were designed and trained on MRI and amyloid PET data that were pre-processed to be as fairly compared as possible. The classification results indicate that MRI data is more conducive to neural network training than amyloid PET data to predict clinical diagnosis. However, a network that uses both modalities, even with the same number of trainable weights, will achieve higher accuracy. This indicates that the two data types have complementary information that can be leveraged in these kinds of tasks.

While these results are a step forward in the optimization of computer aided diagnosis tools for AD, the value from this investigation must be utilized in further applications. To begin, the efficacy of these algorithms should be examined when the MCI state is included in classification. Following this, a natural extension can be made to looking at AD patients on a functional spectrum rather than distinct diagnosis categories. Additionally, as previously alluded to, longitudinal studies that use several scanning sessions of multiple modalities may not only improve classification performance, but also allow the ability to perform more complex tasks such as predicting future cognitive decline irrespective of clinical phenotype. These results would be invaluable to clinicians, as they can directly inform decisions regarding preemptive or preventative care.

\section*{Acknowledgment}
The authors would like to thank the Integrated Data Driven Discovery in Earth and Astrophysical Sciences (IDEAS) program at Northwestern University (NSF Research Traineeship Grant 1450006), the Biomedical Data Driven Discovery (BD3) training program at Northwestern University (NIH Grant 5T32LM012203-02), and the National Institute on Aging (T32AG020506) for financial support. The authors would also like to thank the QUEST High Performance Computing Cluster at Northwestern University for computational resources.

Data collection and sharing for this project was funded by the Alzheimer's Disease Neuroimaging Initiative (ADNI) (National Institutes of Health Grant U01 AG024904) and DOD ADNI (Department of Defense award number W81XWH-12-2-0012). ADNI is funded by the National Institute on Aging, the National Institute of Biomedical Imaging and Bioengineering, and through generous contributions from the following: AbbVie, Alzheimer's Association; Alzheimer's Drug Discovery Foundation; Araclon Biotech; BioClinica, Inc.; Biogen; Bristol-Myers Squibb Company; CereSpir, Inc.; Cogstate; Eisai Inc.; Elan Pharmaceuticals, Inc.; Eli Lilly and Company; EuroImmun; F. Hoffmann-La Roche Ltd and its affiliated company Genentech, Inc.; Fujirebio; GE Healthcare; IXICO Ltd.; Janssen Alzheimer Immunotherapy Research \& Development, LLC.; Johnson \& Johnson Pharmaceutical Research \& Development LLC.; Lumosity; Lundbeck; Merck \& Co., Inc.; Meso Scale Diagnostics, LLC.; NeuroRx Research; Neurotrack Technologies; Novartis Pharmaceuticals Corporation; Pfizer Inc.; Piramal Imaging; Servier; Takeda Pharmaceutical Company; and Transition Therapeutics. The Canadian Institutes of Health Research is providing funds to support ADNI clinical sites in Canada. Private sector contributions are facilitated by the Foundation for the National Institutes of Health (www.fnih.org). The grantee organization is the Northern California Institute for Research and Education, and the study is coordinated by the Alzheimer's Therapeutic Research Institute at the University of Southern California. ADNI data are disseminated by the Laboratory for Neuro Imaging at the University of Southern California.

\bibliographystyle{IEEEtran}
\bibliography{jobname}

\end{document}